\documentclass[11pt,a4paper]{article}
\usepackage[hyperref]{emnlp2018}
\usepackage{times}
\usepackage{latexsym}
\usepackage{graphicx}
\usepackage{url}
\usepackage{verbatim}
\usepackage{amstext}
\aclfinalcopy 

\title{Towards Universal Dialogue State Tracking}

\author{Liliang Ren, Kaige Xie, Lu Chen and Kai Yu \\
   Key Lab. of Shanghai Education Commission for Intelligent Interaction and Cognitive Eng.\\
    SpeechLab, Department of Computer Science and Engineering\\
        Brain Science and  Technology Research Center\\
    Shanghai Jiao Tong University, Shanghai, China\\
  {\tt \{renll204,lightyear0117,chenlusz,kai.yu\}@sjtu.edu.cn} \\}

\date{}

\begin{document}
\maketitle
\begin{abstract}
Dialogue state tracking is the core part of a spoken dialogue system. It estimates the beliefs of possible user's goals at every dialogue turn. However, for most current approaches, it's difficult to scale to large dialogue domains. They have one or more of following limitations: (a) Some models don't work in the situation where slot values in ontology changes dynamically; (b) The number of model parameters is proportional to the number of slots; (c) Some models extract features based on hand-crafted lexicons. To tackle these challenges, we propose StateNet, a {\em universal} dialogue state tracker. It is independent of the number of values, shares parameters across all slots, and uses pre-trained word vectors instead of explicit semantic dictionaries.
Our experiments on two datasets show that our approach not only overcomes the limitations, but also significantly outperforms the performance of state-of-the-art approaches.
\end{abstract}

\section{Introduction}

\begin{figure*}[htb]
\centering
\includegraphics[width=12cm]{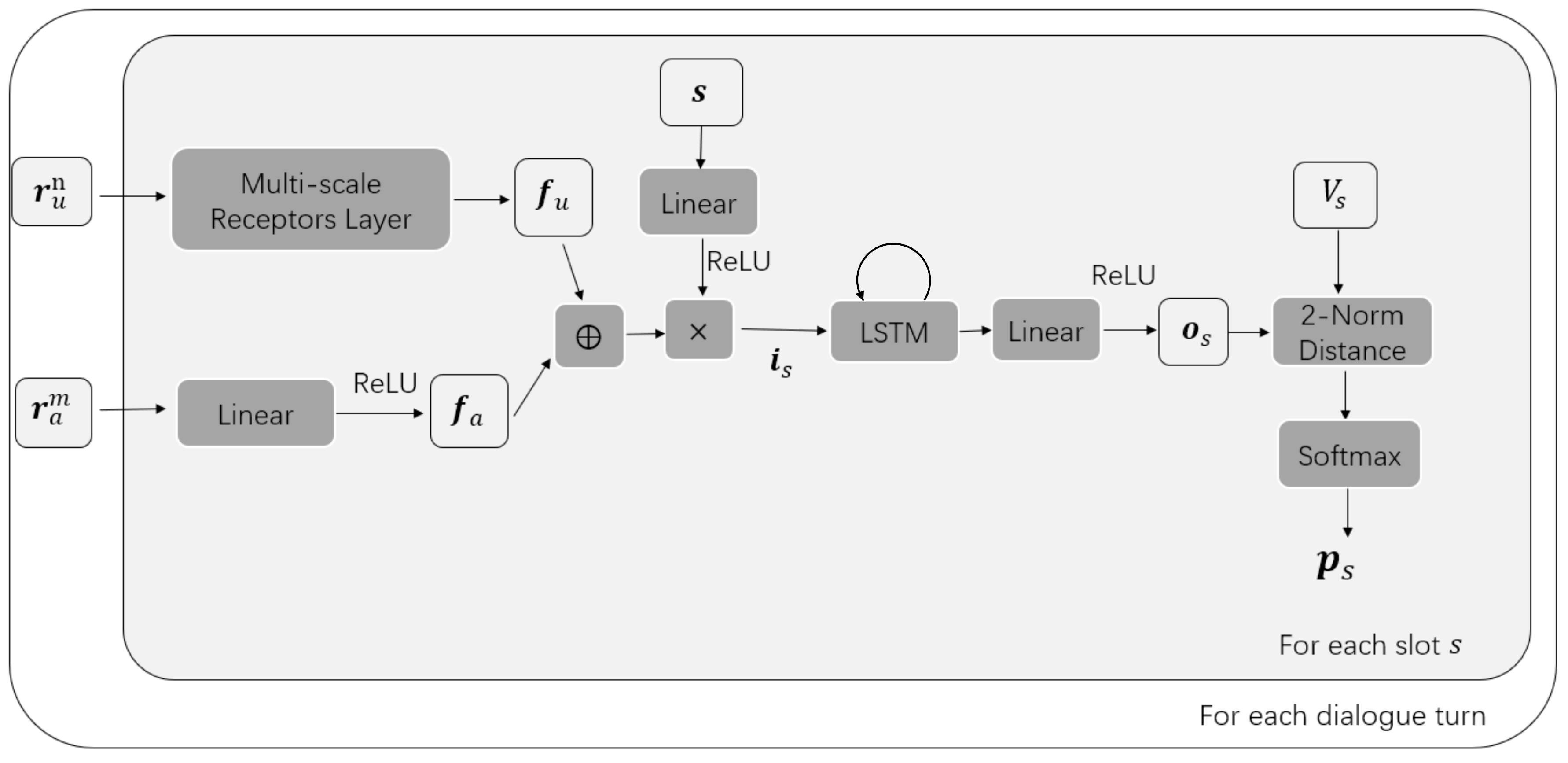}
\caption{General model architecture of StateNet.}\label{f1}
\end{figure*}

A task-oriented spoken dialogue system (SDS) is a system that can continuously interact with a human to accomplish a predefined task through speech. It usually consists of three modules: input, output, and control. The control module is also referred to as {\em dialogue management} \cite{young2010hidden,yu2014cognitive}. It has two missions: dialogue state tracking (DST) and decision making. At each dialogue turn, a state tracker maintains the internal state of the system based on the information received from the input module. Then a machine action is chosen based on the dialogue state according to a dialogue policy to direct the dialogue \cite{chen2018structured}. 

The dialogue state is an encoding of the machine's understanding of the whole conversation. Traditionally, it is usually factorized into three distinct components \cite{young-prcieee13}: the user's {\em goal}, the user's action, and the dialogue history. Among them, the user's goal is most important, which is often simply represented by {\em slot-value} pairs. In this paper, we focus on the tracking of the user's goal.

Recently, the dialogue state tracking challenges (DSTCs) \cite{williams-EtAl:2013:SIGDIAL,henderson-thomson-williams:2014:W14-43,henderson-slt14.1} are organized to provide shared tasks for comparing DST algorithms. A various of models are proposed, e.g.~rule-based models \cite{wang-lemon:2013:SIGDIAL,sun-slt14,yu-tasl15,ky219-yu-fcs15,ks001-sun-is16},  generative statistical models \cite{thomson2010bayesian,young2010hidden,young-prcieee13}, and discriminative statistical models \cite{lee-eskenazi:2013:SIGDIAL,lee:2013:SIGDIAL,sun-EtAl:2014:W14-43,xie-sigdial15,ks001-sun-d&d16,kgx17-xie-sigdial18}. And the state-of-the-art one is the deep learning-based approach. However, most of these models have some limitations. First, some models can only work on a fixed domain {\em ontology}, i.e.~the slots and values are defined in advance, and can't change dynamically. However, this is not flexible in practice \cite{xu2018acl}. For example, in the tourist information domain, new restaurants or hotels are often added, which results in the change of the ontology. Second, in many approaches the models for every slot are different. Therefore, the number of parameters is proportional to the number of slots. Third, some models extract features based on text {\em delexicalisation} \cite{henderson-slt14.2}, which depends on predefined semantic dictionaries. In large scale domains, it's hard to manually construct the semantic dictionaries for all slots and values \cite{mrkvsic2017neural}.

To tackle these challenges, here we propose a {\em universal} dialogue state tracker, StateNet. For each state slot, StateNet generates a fixed-length representation of the dialogue history, and then compares the distances between this representation and the value vectors in the candidate set for making prediction. The set of candidate values can change dynamically. StateNet only needs the following three parts of the data: (1) the original ASR information (or the transcript) of the user utterance; (2) the information of the machine act; (3) the literal names of the slots and the values. The manually-tagging of the user utterance is not needed as a part of the data. StateNet shares parameters among all slots, through which we can not only transfer knowledge among slots but also reduce the number of parameters.

\section{StateNet: A Universal Dialogue State Tracker}
\label{sec:modelStructure}
For each dialogue turn, StateNet takes the multiple $n$-gram user utterance representation, $\mathbf{r}_{u}^n$, the $m$-gram machine act representation, $\mathbf{r}_{a}^m$, the value set, $\mathcal{V}_{s}$, and the word vector of the slot, $\mathbf{s}$, as the input. Then StateNet applies the Long Short-Term Memory (LSTM) \cite{hochreiter1997long} to track the inner dialogue states among the dialogue turns. And for each slot, StateNet outputs a corresponding probability distribution, $\mathbf{p}_{s}$, over the set of possible values, $\mathcal{V}_{s}$, at each of the dialogue turn,
\[
\mathbf{p}_{s}=\text{StateNet}(\mathbf{r}_{u}^n,\mathbf{r}_{a}^m,\mathbf{s},\mathcal{V}_{s}).
\label{eq0}
\]

The general model architecture is shown in Figure~\ref{f1}.

\subsection{User Utterance Representation}

At the $t$-th dialogue turn, the user utterance, $U_t$, may consist of $l$ number of words, $u_{i}$, with their corresponding word vectors, $\mathbf{u}_{i}$, ($1 \leq i \leq l$). The user utterance may also have its corresponding $m$-best ASR hypotheses with the normalized confidence scores \cite{zhc00-chen-icassp17}, $q_j$,($1 \leq j \leq m$). In this case, we can calculate the weighted word vectors, $\mathbf{u'}_{i}$,
\[
\mathbf{u'}_{i}=\sum_{j=1}^{m} q_j \mathbf{u}_{i,j},
\vspace{-2mm}
\]
where $\mathbf{u}_{i,j}$ represents the word vector $\mathbf{u}_{i}$ presented at the $j$-th ASR hypothesis, and the zero vectors are padded at the end of all the hypotheses that are shorter than the longest one to have a same length of the utterance.

Based on the weighted word vectors generalizing the information from the ASR hypothesis, we can then construct the $n$-gram weighted word vectors, as proposed by \citeauthor{mrkvsic2017neural} \shortcite{mrkvsic2017neural}, 
\[
\mathbf{u'}_{i}^{n}=\mathbf{u'}_{i}\oplus...\oplus\mathbf{u'}_{i+n-1},
\label{eq2}
\]
where $\oplus$ is the concatenation operator between the word vectors.

An $n$-gram user utterance representation is then constructed through a sum of the $n$-gram weighted word vectors,
\[
\mathbf{r}_{u}^n=\sum_{i=1}^{l-n+1} \mathbf{u'}_{i}^{n}.
\label{eq2}
\]

\subsection{Multi-scale Receptors Layer}

\begin{figure}[htb]
\centering
\includegraphics[width=7.8cm]{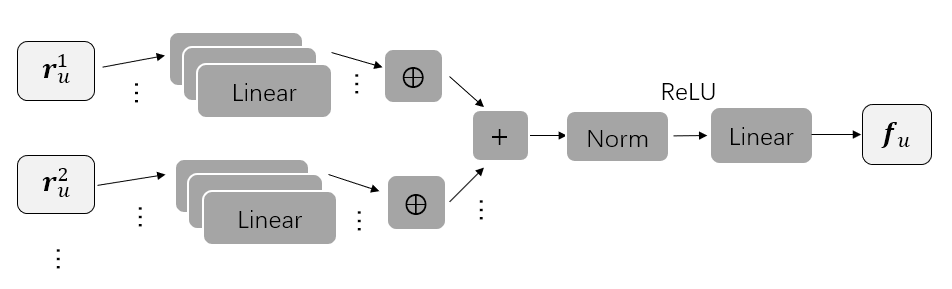}
\caption{Multi-scale Receptors Layer.}\label{f2}
\end{figure}
For each gram $k$ of the user utterance representation, $\mathbf{r}_{u}^k$, ($1 \leq k \leq n$), the Multi-scale Receptors Layer has $c$ number of linear neural networks (with the same number of neurons, $N_c$). Each of them takes the representation as input and is expected to work as the specialized receptor to amplify the signals from some of the word vectors in the utterance representation,
\[
\mathbf{\hat{r}}_{u}^k=\oplus_{j=1}^{c}(\mathbf{W}_{k}^{j}\mathbf{r}_{u}^{k}+\mathbf{b}_{k}^{j}),      
\label{eq3}
\]
where $\mathbf{W}_{k}^{j}$ means the weight of the $j$-th linear layer, $\mathbf{b}_{k}^{j}$ means the corresponding bias, and $\oplus$ is the concatenation operator between the neurons of these linear layers. Note that each receptor does not necessarily has to be a single linear neural network and can be sophisticated with multiple layers and non-linearity for better detection performance. Here we only use the linear layer to provide a baseline of this kind of structure design.

These $c$ number of linear layers (or receptors) for different grams (or scales) of the representation $\mathbf{\hat{r}}_{u}^k$ is then summed together to be layer-normalized \cite{ba2016layer}. After that, the ReLU activation function is applied, followed by a linear layer with the size $N_c$ that maps all the receptors to a user feature vector, $\mathbf{f}_{u}$,
\[
\mathbf{f}_{u}=\text{Linear}(\text{ReLU}(\text{LayerNorm}(\sum_{k=1}^{n}\mathbf{\hat{r}}^k_{u}))).
\label{eq3}
\]

\subsection{Machine Act Representation}

We represent the machine act in the $m$ order n-gram of bag of words, $\mathbf{r}_a^m$, based on the vocabularies generalized from the machine acts in the training set of a given data set. The machine act feature, $\mathbf{f}_{a}$, is then simply generated through a linear layer of size $N_c$ with the ReLU activation function,
\[
\mathbf{f}_{a}=\text{ReLU}(\text{Linear}(\mathbf{r}_a^m)).
\label{eq3}
\]

\subsection{Slot Information Decoding}
Since a slot, e.g.~{\em area} or {\em food}, is usually indicated as a word or a short word group, then it can be represented as a single word vector (with multiple word vectors summed together), $\mathbf{s}$. A single linear layer with the size $2N_c$ is applied to the word vector $\mathbf{s}$, followed by the ReLU non-linear layer,
\[
\mathbf{f}_{s}=\text{ReLU}(\text{Linear}(\mathbf{s})).
\label{eq3}
\]

The turn-level feature vector, $\mathbf{i}_{s}$, is then generated through a point-wise multiplication $\otimes$ between the slot feature and the concatenation of the user feature and the machine act feature,
\[
\mathbf{i}_{s}=\mathbf{f}_{s}\otimes(\mathbf{f}_{u}\oplus\mathbf{f}_{a}).
\label{eq3}
\]

In this way, the turn-level feature vector is intended to amplify the large magnitude signals that are from both the user and machine act feature vector and the slot feature vector.

\subsection{Fixed-length Value Prediction}
Given the turn-level feature vector, $\mathbf{i}_s$, we can now track the dialogue state throughout the dialogue turns by LSTM.  For the current turn $t$, the LSTM takes the $\mathbf{i}_s$ and the previous hidden state, $\mathbf{q}_{t-1}$, as the input. We can then obtain a fixed-length value prediction vector, $\mathbf{o}_{s}$, whose length is equal to $N_w$, i.e.~the dimension of the word vectors which are fed into the model,
\[
\mathbf{o}_{s}= \text{ReLU}(\text{Linear}(\text{LSTM}(\mathbf{i}_{s},\mathbf{q}_{t-1}))),
\label{eq3}
\]
where the linear layer has $N_w$ neurons. In this way, the prediction of the model is independent of the number of the given values, so it is possible for the model to perform parameter sharing among each of the slots. The fixed-length prediction can somehow be interpreted as a \emph{word vector} that is ready for the calculation of the similarity between the prediction and the true value label.

\subsection{2-Norm Distance}

For a specific semantic slot, since there may be no corresponding value in a given dialogue turn, thus we always add a literally ``none'' value to the value set for the model to track this state. For the evaluation of the similarity between the prediction and the value, we calculate the 2-Norm distance between the prediction vector and each of the word vectors of the values in the value set. Softmax function is performed with respect to all the negative relative distances to give a distribution of probabilities for the values, $v_i \in \mathcal{V}_{s}$,
\[
    p_s(v_i) = \text{Softmax}(-||\mathbf{o}_{s}-\mathbf{v}_i||),
\]
where $\mathbf{v}_i$ is the representation vector of $v_i$. If the slot value $v_i$ consists of more than one word, $\mathbf{v}_i$ will then be the summation of all corresponding word vectors.
When training the model, we minimize the Cross-Entropy (CE) loss between the output probabilities and the given label.

StateNet requires the user utterance, the semantic slots, and slot values to be able to be expressed in words and have their corresponding word vectors. We use the fixed word embedding for every word, and do not fine-tune the word embeddings in the model. Since the word embeddings are distributed on a fixed-dimension vector space and hold rich semantic information, StateNet may have the ability to track the dialogue state for any new slot or value, as long as the corresponding word embedding can be found. This is the reason why we call the StateNet a {\em universal} dialogue state tracker.


\section{Experiments}

\begin{table*}[t]
  \centering
  \begin{tabular}{|c|c|c|}
  \hline
    \textbf{DST Models} & \begin{tabular}[c]{@{}c@{}}\textbf{Joint Acc.}\\\textbf{DSTC2}\end{tabular} & \begin{tabular}[c]{@{}c@{}}\textbf{Joint Acc.}\\\textbf{WOZ 2.0}\end{tabular} \\ 
    \hline
    Delexicalisation-Based (DB) Model \cite{mrkvsic2017neural} & 69.1 & 70.8 \\ 
    DB Model + Semantic Dictionary \cite{mrkvsic2017neural} & 72.9 & 83.7 \\
    Scalable Multi-domain DST \cite{rastogi2017scalable} & 70.3 & - \\
    \hline
    MemN2N \cite{perez2017dialog} & 74.0 & - \\
    PtrNet \cite{xu2018acl} & 72.1 & - \\
    Neural Belief Tracker: NBT-DNN \cite{mrkvsic2017neural} & 72.6 & 84.4 \\
    Neural Belief Tracker: NBT-CNN \cite{mrkvsic2017neural} & 73.4 & 84.2 \\
    Belief Tracking: Bi-LSTM \cite{ramadan2018large} & - & 85.1 \\
    Belief Tracking: CNN \cite{ramadan2018large} & - & 85.5 \\
    GLAD \cite{zhong2018global} & 74.5 & 88.1 \\
    \hline
    StateNet & 74.1 & 87.8 \\
    StateNet\_PS & 74.5 & 88.2 \\
    \textbf{StateNet\_PSI} & \textbf{75.5} & \textbf{88.9} \\
    \hline
  \end{tabular}
  \vspace{1mm}
  \caption{Joint goal accuracy on DSTC2 and WOZ 2.0 test set vs.~various approaches as reported in the literature.}
  \label{tab:res}
\end{table*}

Experiments are conducted to assess the performance on joint goal. Two datasets are used by us for training and evaluation. One is the second Dialogue State Tracking Challenge (\textbf{DSTC2}) dataset~\cite{henderson-thomson-williams:2014:W14-43}, and the other is the second version of Wizard-of-Oz (\textbf{WOZ 2.0}) dataset~\cite{wenN2N17}. Both of them are the conversations between users and a machine system. The user's goal is to find a suitable restaurant around Cambridge. The ontology of these two datasets is identical, which is composed of three informable slots: \emph{food, pricerange} and \emph{area}. The main difference between them is that in WOZ 2.0, users typed instead of using speech directly. This means the users can use far more sophisticated language than they can in the DSTC2, which is a big challenge for the language understanding ability of the model. Thus, it allows WOZ 2.0 to be more indicative of the model's actual performance since it is immune to ASR errors. 

Based on the model structure as described in Section \ref{sec:modelStructure}, we implement three kinds of dialogue state tracker. The difference among them lies in the utilization of parameter sharing and parameter initialization.

\begin{itemize}
    \setlength{\abovedisplayskip}{0pt}
    \setlength{\belowdisplayskip}{0pt}
    \setlength{\itemsep}{0pt}
    \setlength{\parskip}{0pt}
    \item \texttt{StateNet}: It doesn't have shared parameters among different slots. In other words, three models for three slots are trained separately using RMSProp optimizer, learning rate set to 0.0005. And its parameters are not initialized with any pre-trained model.
    \item \texttt{StateNet\_PS}: Parameter sharing is conducted among three slots. For each slot in a batch, we infer the model with the slot information and the same dialogue information. The losses are calculated based on the corresponding value set. After each slot is inferred, we back-propagate all the losses and do the optimization. So we just train one model in total using RMSProp optimizer, learning rate set to 0.0005. As a result, the amount of model parameters is one third of that of \texttt{StateNet}, which means \texttt{StateNet\_PS} can significantly save the memory usage during inferring.
    \item \texttt{StateNet\_PSI}: Parameter sharing is conducted within this model, same as \texttt{StateNet\_PS}, but its parameters are initialized with a pre-trained model. For pre-training, we only allow the model to track one single slot and make predictions on its value set. After the training ends, we save the model parameters and use them to initialize the model parameters for the training of the multi-slot tracking. The pre-trained model with the best performance on the validation set is selected for initialization. Here, we choose the \emph{food} slot for pre-training since \texttt{StateNet} has the lowest prediction accuracy on the \emph{food} slot. \texttt{StateNet\_PSI} is trained using Adam optimizer and learning rate is set to 0.001. Since the model has obtained the basic knowledge from the pre-trained model, then a more aggressive learning process is preferred. Adam with a higher learning rate can help a lot compared to RMSProp optimizer.
\end{itemize}

The hyperparameters are identical for all three models, $N_c=128,N_w=300,n=2,m=3$. We use $c=4$ for the number of the receptors for each slot, where the number is determined through the grid search. The word embeddings used by us is the \emph{semantically specialised} Paragram-SL999 vectors \cite{wieting2015paraphrase} with the dimension of 300, which contain richer semantic contents compared to other kinds of word embeddings. Implemented with the MXNet deep learning framework of Version 1.1.0, the model is trained with a batch size of 32 for 150 epochs on a single NVIDIA GTX 1080Ti GPU.

The results in Table \ref{tab:res} show the effectiveness of parameter sharing and initialization. \texttt{StateNet\_PS} outperforms \texttt{StateNet}, and \texttt{StateNet\_PSI} performs best among all 3 models. It is because the parameter sharing can not only prevent the model diverging from the right learning process but also transfer necessary knowledge among different slots. And the parameter initialization provides the model with the opportunity to gain some basic while essential semantic information at the very beginning since the \emph{food} slot is the most important and difficult one. Besides, \texttt{StateNet\_PSI} beats all the models reported in the previous literature, whether the model with delexicalisation \cite{henderson-slt14.2,henderson-thomson-young:2014:W14-43,rastogi2017scalable} or not \cite{mrkvsic2017neural,perez2017dialog,xu2018acl,ramadan2018large,zhong2018global}.

\begin{table}[htb]
  \centering
  \begin{tabular}{|c|c|c|}
  \hline
    \textbf{Initialization} & \begin{tabular}[c]{@{}c@{}}\textbf{Joint Acc.}\\\textbf{DSTC2}\end{tabular} & \begin{tabular}[c]{@{}c@{}}\textbf{Joint Acc.}\\\textbf{WOZ 2.0}\end{tabular} \\ 
    \hline
     \emph{food} & \textbf{75.5} & \textbf{88.9} \\ 
     \emph{pricerange} & 73.6 & 88.2 \\
     \emph{area} & 73.5 & 87.8 \\
    \hline
  \end{tabular}
\vspace{1mm}
  \caption{Joint goal accuracy on DSTC2 and WOZ 2.0 of \texttt{StateNet\_PSI} using different pre-trained models based on different single slot.}
  \label{tab:ini}
\end{table}

We also test \texttt{StateNet\_PSI} with different pre-trained models, as shown in Table \ref{tab:ini}. The fact that the \emph{food} initialization has the best performance verifies our selection of the slot with the worst performance for pre-training. This is because the good performance on joint goal requires a model to make correct predictions on all of the slots. A slot on which the model has the worst accuracy, i.e.~the most difficult slot, will dramatically limit the overall model performance on the metric of the joint goal accuracy. Thus, the initialization with a model pre-trained on the most difficult slot can improve the performance of the model on its weakness slot and boost the joint goal accuracy, while the initialization of a strength slot may not help much for the overall accuracy but in turn causes the over-fitting problem of the slot itself.

\section{Conclusion}

In this paper, we propose a novel dialogue state tracker that has the state-of-the-art accuracy as well as the following three advantages: 1) the model does not need manually-tagged user utterance; 2) the model is scalable for the slots that need tracking, and the number of the model parameters will not increase as the number of the slots increases, because the model can share parameters among different slots; 3) the model is independent of the number of slot values, which means for a given slot, the model can make the prediction on a new value as long as we have the corresponding word vector of this new value. If there are a great number of values for a certain slot, to reduce the computational complexity, we can utilize a fixed-size candidate set \cite{rastogi2017scalable}, which dynamically changes as the dialogue goes on. Experiment results demonstrate the effectiveness of parameter sharing \& initialization. 

Our future work is to evaluate the performance of our models in the scenario where there are new slots and more unobserved slot values, and to evaluate the domain-transferring ability of our models.

\section*{Acknowledgments}

The corresponding author is Kai Yu.~This work has been supported by the National Key Research and Development Program of China (Grant~No.~2017YFB1002102) and the China NSFC project (No.~61573241). Experiments have been carried out on the PI supercomputer at Shanghai Jiao Tong University.


\bibliography{ref}
\bibliographystyle{acl_natbib_nourl}

\end{document}